\documentclass[lettersize,journal]{IEEEtran}
\usepackage{amsmath,amsfonts}
\usepackage{algorithmic}
\usepackage{algorithm}
\usepackage{array}
\usepackage[caption=false,font=normalsize,labelfont=sf,textfont=sf]{subfig}
\usepackage{textcomp}
\usepackage{stfloats}
\usepackage{url}
\usepackage{verbatim}
\usepackage{graphicx}
\usepackage{cite}
\usepackage{array}
\usepackage{multirow}
\usepackage{colortbl}
\usepackage[table]{xcolor}
\usepackage{marvosym}
\usepackage{fix-cm}
\usepackage{textcomp}
\usepackage{hyperref}
\usepackage{booktabs, multirow, xcolor}

\newcommand{\equalcontrib}{\textsuperscript{*}}
\newcommand{\corresponding}{\textsuperscript{\dag}}

\begin{document}

\title{
% RT-OVAD: Open Vocabulary Aerial Object Detection \\ with Image-Text Collaboration
RT-OVAD: Real-Time Open-Vocabulary Aerial Object Detection \\ via Image-Text Collaboration
}
% RT-OVA: Real-Time Open-Vocabulary Aerial Object Detection via Image-Text Collaboration

\author{
    Guoting Wei\equalcontrib, 
    Xia Yuan\equalcontrib, 
    Yu Liu\equalcontrib, 
    Zhenhao Shang, 
    Xizhe Xue,
    Peng Wang,
    Kelu Yao, \\
    Chunxia Zhao,
    Haokui Zhang\corresponding,
    and Rong Xiao
        % <-this % stops a space
\thanks{This work was done during Guoting Wei’s internship at IntelliFusion. \equalcontrib represents equal contribution. \corresponding represents the corresponding author.}% <-this % stops a space
% \thanks{G.~Wei, X.~Yuan, and C.~Zhao are with Nanjing University of Science and Technology, Nanjing 210094, China (e-mail: \{weiguoting, yuanxia, zhaochx\}@njust.edu.cn).}%
% %
% \thanks{Z.~Shang, X.~Xue, P.~Wang, and H.~Zhang are with Northwestern Polytechnical University, Xi’an 710072, China (e-mail: \{shangzhenhao, xuexizhe\}@mail.nwpu.edu.cn, \{peng.wang, hkzhang\}@nwpu.edu.cn).}%
% %,
% \thanks{Y.~Liu and K.~Yao are with Zhejiang Lab, Hangzhou 311121, China (e-mail: \{UniLiu, yaokelu\}@zhejianglab.org).}%
% %
% \thanks{R.~Xiao is with Intellifusion Inc., Shenzhen 518000, China (e-mail: xiao.rong@intellif.com).}%
% }
\thanks{G.~Wei, X.~Yuan, and C.~Zhao are with Nanjing University of Science and Technology, Nanjing 210094, China (e-mail: \{weiguoting, yuanxia, zhaochx\}@njust.edu.cn).}%
\thanks{Z.~Shang, X.~Xue, P.~Wang, and H.~Zhang are with Northwestern Polytechnical University, Xi'an 710072, China (e-mail: \{shangzhenhao, xuexizhe\}@mail.nwpu.edu.cn; \{peng.wang, hkzhang\}@nwpu.edu.cn).}%
\thanks{Y.~Liu and K.~Yao are with Zhejiang Lab, Hangzhou 311121, China (e-mail: \{UniLiu, yaokelu\}@zhejianglab.org).}%
\thanks{H.~Zhang and R.~Xiao are with Intellifusion Inc., Shenzhen 518000, China (e-mail: rongxiao@gmail.com).}%
}
\markboth{IEEE Transactions on Image Processing,~Vol.~XX, No.~X, Month~Year}%
{Author \MakeLowercase{\textit{et al.}}: Title}


\maketitle

\begin{abstract}
Aerial object detection plays a crucial role in numerous applications. However, most existing methods focus on detecting predefined object categories, limiting their applicability in real-world open scenarios. In this paper, we extend aerial object detection to open scenarios through image-text collaboration and propose RT-OVAD, the first real-time open-vocabulary detector for aerial scenes. 
Specifically, we first introduce an image-to-text alignment loss to replace the conventional category regression loss, thereby eliminating category constraints. 
Next, we propose a lightweight image–text collaboration strategy comprising an image–text collaboration encoder and a text-guided decoder. The encoder simultaneously enhances visual features and refines textual embeddings, while the decoder guides object queries to focus on class-relevant image features. This design further improves detection accuracy without incurring significant computational overhead.
Extensive experiments demonstrate that RT-OVAD consistently outperforms existing state-of-the-art methods across open-vocabulary, zero-shot, and traditional closed-set detection tasks. For instance, on the open-vocabulary aerial detection benchmarks DIOR, DOTA-v2.0, and LAE-80C, RT-OVAD achieves 87.7 AP$_{50}$, 53.8 mAP, and 23.7 mAP, respectively, surpassing the previous state-of-the-art (LAE-DINO) by 2.2, 7.0, and 3.5 points. In addition, RT-OVAD achieves an inference speed of 34 FPS on an RTX 4090 GPU, approximately three times faster than LAE-DINO (10 FPS), meeting the real-time detection requirements of diverse applications. The code will be released at \url{https://github.com/GT-Wei/RT-OVAD}.

\end{abstract}


\section{Introduction}
Aerial object detection, which involves localizing and categorizing objects of interest within aerial images, has become increasingly important due to its broad application requirements, including earth monitoring, disaster search, and rescue \cite{li2020object, ding2021object, sadgrove2018real}.
Most existing aerial detectors \cite{du2023adaptive, yang2019clustered, li2020density, huang2022ufpmp, li2021gsdet, 10767172, meethal2023cascaded} can only identify predefined categories from their training sets, which  strictly restrict their deployment in open-world scenarios, which have become key considerations for real-world applications. A major obstacle for these detectors in discovering novel classes is their reliance on mapping image features to a fixed set of class indices\cite{han2023s}. 

\begin{figure}[!t]
\centering
\includegraphics[width=1.0\linewidth]{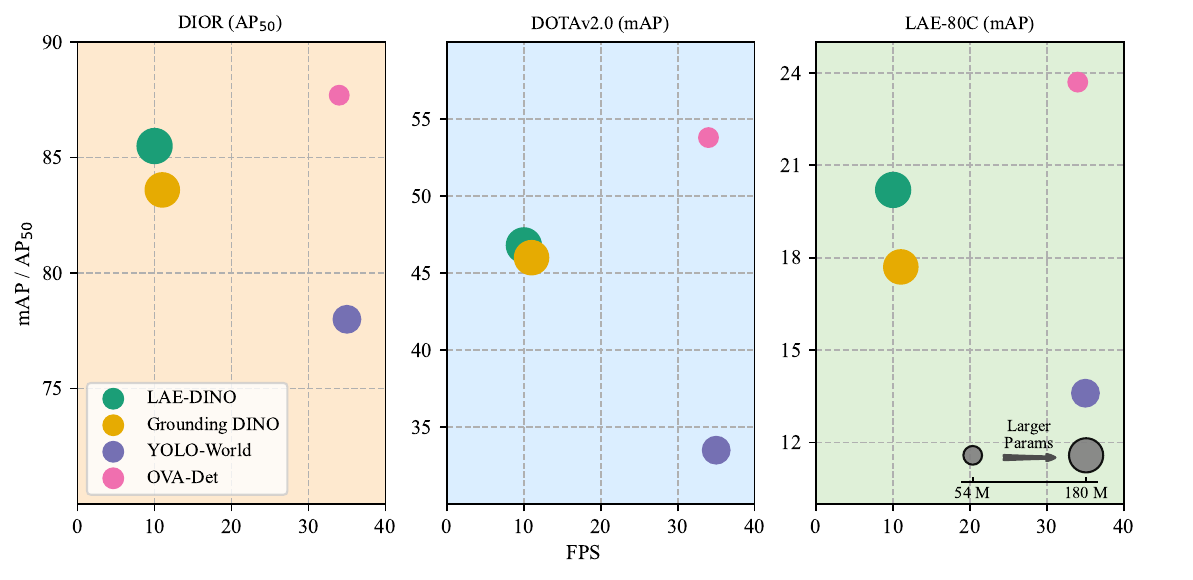}
\caption{
% Comparison of RT-OVAD with other advanced open-vocabulary detectors on the DIOR, DOTAv2.0, and LAE-80C datasets in terms of mAP/AP$_{50}$, FPS, and Param. Marker size encodes Param, such that biger markers indicate larger params. All models are evaluated in open-vocabualry aerial detection setting.
Comparison of RT-OVAD with other advanced open-vocabulary detectors on the DIOR, DOTA-v2.0, and LAE-80C datasets in terms of mAP/AP$_{50}$, FPS, and parameter count. Marker size indicates the number of parameters, with larger markers representing models with more parameters. All models are evaluated under the open-vocabulary aerial detection setting.
}
\label{fig:comparison}
\end{figure}

Recently, CLIP \cite{radford2021learning} has the capability to map vision-language features into a comparable representation space. This offers an opportunity to break the category limitations inherent in traditional object detection. Inspired by this, several works \cite{gu2021open, zhao2022exploiting, liu2024grounding, cheng2024yolo} on natural images have explored the relationship between images and text to endow models with open-vocabulary detection capabilities, achieving notable success. For instance, VILD \cite{gu2021open} and VL-PLM \cite{zhao2022exploiting} use pre-trained VLM to classify the object proposals generated by region proposal network. GLIP \cite{li2022grounded} formulates object detection as a phrase grouding task, thereby unifying these two tasks. The introduction of textual information and the synergy between image and text have significantly expanded the coverage of traditional object detection tasks, taking them a step closer to real-world applications in open environments. 

% However, in the field of aerial object detection, such work remains very limited. Aerial object detection poses unique challenges, including predominantly small objects, varying viewpoints, and limited detection data. Directly applying existing open-vocabulary detection methods designed for conventional perspectives to aerial perspectives may result in suboptimal performance. Very recently, some researchers have attempted to address this gap and proposed insightful methods \cite{zang2024zero, li2024toward, li2024exploiting, pan2025locate, huang2025openrsd}. The introduction of these works has validated the potential of the open-vocabulary approach in the aerial object detection field, but there is still significant room for optimization. 
% First, image-text alignment is significant challenge for open-vocabulary detection, existing methods directly adopts the previous alignment方法 in 自然图像检测器中, which很容易过拟合在有限的detection数据中. Secondly, current open-vocabulary aerial neglect the importance of visual-semantic interaction, especially the background interference in this process. Finally, there is a paucity of efficient end-to-end open-vocabulary detectors. Most exsiting detectors adopt 复杂的半监督训练策略, this stragegy need 借助构建伪标签来识别新类. Although LAE-DINO as a end-to-end detectors build on Grounding DINO, 但它neglects the efficiency.

However, in the field of aerial object detection, such work remains limited. Aerial object detection poses unique challenges, including predominantly small objects, varying viewpoints, background interference, and limited detection data. Directly applying existing open-vocabulary detection methods designed for conventional perspectives to aerial perspectives may result in suboptimal performance. Very recently, some researchers have attempted to address this gap and proposed insightful methods \cite{zang2024zero, li2024toward, li2024exploiting, pan2025locate}. These works have demonstrated the potential of the open-vocabulary approach in the aerial object detection field, but there is still significant room for optimization.
First, image-text alignment poses a significant challenge for open-vocabulary aerial detection. Existing methods often adopt alignment strategies developed for natural-image-based detectors, which may lead to overfitting due to the limited aerial detection data.
Secondly, current open-vocabulary aerial detectors typically overlook visual-semantic interaction and, in particular, fail to address the background interference that arises during this process, which is especially prominent in aerial imagery.
Finally, there remains a scarcity of efficient end-to-end open-vocabulary aerial detectors. Most existing methods \cite{li2024toward, li2024exploiting} utilize complex semi-supervised training strategies that heavily rely on pseudo-label generation for discovering novel classes. Although LAE-DINO \cite{pan2025locate} achieves an end-to-end framework based on Grounding DINO, it still falls short in achieving real-time inference efficiency.

% For instance, DescReg \cite{zang2024zero} introduces a description regularization method that focuses on enhancing textual descriptions but overlooks the importance of visual-semantic interaction. CastDet \cite{li2024toward} adopts a complex multi-teacher framework based on a two-stage detector for open-vocabulary detection, which neglects the efficiency. 
% LAE-DINO \cite{pan2025locate} utilizes vision-language models to expand detectable categories and employs visually-guided text prompt learning, but it neglects significant background interference during visual-textual interactions, especially prevalent in aerial images.

In this paper, we propose a real-time open-vocabulary detection model specifically designed for aerial photography scenes, termed RT-OVAD. Specifically, we adopt the basic architecture of RT-DETR, due to its high inference efficiency and concise structure. 
First, we adapt the pretrained CLIP model and integrate it into the base framework to leverage its image-text alignment capability and powerful feature extraction, alleviating the limited sample problem. We then propose an image-to-text alignment loss to bridge the domain gap between natural images and aerial images, extending the model's applicability from closed-world to open-world scenarios. 
Next, considering the challenges posed by long perception distances in aerial images, such as small detection targets with less prominent visual features, strong background interference, and intra-class instance scale variations, we propose lightweight image–text collaboration architectures consisting of an image–text collaboration encoder and a text-guided decoder. The encoder enhances both visual and textual representations, while the decoder guides object queries to focus on class-relevant features. This design further improves detection accuracy without incurring substantial computational costs.
Finally, we conduct comprehensive experiments across three benchmark tasks: open-vocabulary detection, zero-shot detection, and traditional closed-set detection. As illustrated in Figure~\ref{fig:comparison}, RT-OVAD consistently surpasses mainstream open-vocabulary detectors (GLIP, Grounding DINO, YOLO-World) as well as the recent aerial-specific detector LAE-DINO in terms of both mAP and inference speed, demonstrating the effectiveness of the proposed approach.

% a lightweight image-text collaboration strategy to further improve detection accuracy without substantially increasing computational costs. This strategy  and  and a text-guided decoder. The image–text collaboration encoder aim to enhances visual features 

% driven by two complementary ideas: text-guided visual feature enhancement, which enhances visual features and object queries in both encoder and decoder, and visual-guided text refinement, which enriches textual embeddings with visual cues from corresponding images to better align class semantics with visual features and mitigates the impact of significant intra-class variations arising from viewpoint and scale differences in aerial imagery.

% which enriches textual embeddings with visual cues from corresponding images to better align class semantics with visual features and mitigates the impact of significant intra-class variations arising from viewpoint and scale differences in aerial imagery. Besides, the proposed image-text collaboration strategy further enhances detection accuracy without substantially increasing computational costs.

In summary, there are three major contributions:
\begin{enumerate}
    \renewcommand{\labelenumi}{\textbullet}
    \item We propose an image-to-text alignment loss as a bridge to address the modality gap between natural images and aerial images. By integrating it with CLIP, we extend the RT-DETR framework from closed-world to open-world scenarios. 
    \item 
    We introduce lightweight image–text collaboration architectures that enhances both visual and textual representations while suppressing background interference, further improving detection accuracy. This strategy comprises three specific modules: Text-Guided Feature Enhancement (TG-FE) and Visual-Guided Text Refinement (VG-TR), which operate in the encoder, and Text-Guided Query Enhancement (TG-QE), which operates in the decoder.
   \item Our method achieves SOTA performance across multiple benchmark datasets for open-vocabulary and zero-shot aerial detection, as well as traditional closed-set tasks, while maintaining 34 FPS. To the best of our knowledge, RT-OVAD is the first real-time open-vocabulary aerial object detection method.
   % \item We introduce a lightweight image–text collaboration strategy to further enhance detection accuracy. This strategy is driven by two core ideas: text-guided visual feature enhancement and visual-guided text refinement. Based on these ideas, we propose three specific modules: Text-Guided Feature Enhancement (TG-FE) and Text-Guided Query Enhancement (TG-QE), which strengthen visual features and suppress background interference, and Visual-Guided Text Refinement (VG-TR), which better to aligns class semantics with visual features and mitigates substantial intra-class variations commonly observed in aerial imagery.

   % For example, RT-OVAD surpasses previous methods by 2.2 AP$_{50}$, 7.0 mAP, and 3.5 mAP on the open-vocabulary aerial benchmarks DIOR, DOTAv2.0, and LAE-80C, respectively, and maintains a speed of 34 FPS. To the best of our knowledge, RT-OVAD is the first real-time open-vocabulary aerial object detection method.
 \end{enumerate}

\section{Related Work}
\label{sec:Related Work}

\subsection{Open Vocabulary Object Detection} 
Traditional object detectors \cite{ren2015faster, cai2018cascade, carion2020end, zhao2024detrs} predominantly rely on fully supervised training, mapping image features to class indices \cite{han2023s}. Such detectors are limited to recognizing only the categories seen during training, known as fixed-vocabulary (closed-set) detection. In contrast, open-vocabulary object detection (OVD) aims to explore the relationship between image and text to enable detectors to identify classes beyond those in the training sets. 
Inspired by CLIP \cite{radford2021learning} establish image-text feature alignment, recent works have incorporated class semantic information into detectors, exploring open-vocabulary object detection. These approaches can be roughly divided into two types. The first type \cite{9318563, gu2021open, zhao2022exploiting, wu2023aligning, li2024learning, zhong2022regionclip, chen2024rtgen} uses networks like FPN to extract object proposals, which are then classified using pre-trained VLMs. For example, VILD \cite{gu2021open} enhances OVD capability by distilling knowledge from pre-trained VLMs. Similarly, VL-PLM \cite{zhao2022exploiting} leverages pre-trained VLMs to generate region-text pair labels for unlabeled images, these labels are subsequently used to train the detector, enabling recognition of novel object categories not present in the original training set.
The second type integrates detection and grounding tasks \cite{li2022grounded, 10247123, zhang2022glipv2, liu2024grounding, yao2023detclipv2, cheng2024yolo}, aiming to collect sufficient data to train specialized VLMs from scratch for detection. For instance, GLIP \cite{li2022grounded} unifies object detection and phrase grounding tasks to pre-train a detector for open-vocabulary detection. Grounding DINO \cite{liu2024grounding} incorporates the pre-training strategy into transformer detection \cite{zhang2022dino} based on phrase grounding. 

The most relevant work is YOLO-World \cite{cheng2024yolo}, which extends YOLO to the open-vocabulary detection field through visual-language modeling. Firstly, YOLO-World and RT-OVAD are proposed for different domains, natural images and aerial images. Secondly, the module design and loss function are also different. Unlike YOLO-World, which primarily uses text as guidance and focuses only on the feature extraction part, our proposed RT-OVAD leverages text information to guide and enhance features in both the encoder and decoder with a concise architecture. This design is more effective in addressing the unique challenges of aerial images, such as low-quality and various viewpoints issues. Furthermore, the image-to-text alignment loss in RT-OVAD replaces bidirectional alignment with unidirectional alignment, reducing optimization difficulty. These differences allow RT-OVAD to achieve better performance in the aerial domain.

% The most relevant work is YOLO-World \cite{cheng2024yolo}, which extend YOLO to open vocabulary detection field through visual language modeling and propose text-guided cross-stage partial layer to improve performance. Differing from YOLO-World which just using text as guidance and focus on feature extraction part only, our proposed RT-OVAD use text information to guide and enhance features in both encoder and decoder part with a concise architecture, which is more useful in addressing the unique challenges in aerial images, such as low-quality issues.  In addition, the image-to-text loss proposed in RT-OVAD reform the bidirectional alignment with unidirectional alignment, reducing the optimization difficulty, making it more suitable for the aerial scene. Benefiting from optimization designs tailored for aerial data, the proposed RT-OVAD outperforms YOLO-World by a large margin, as shown in Table 1 and 2. 

\subsection{Aerial Object Detection}
% In aerial images, objects predominantly appear at small scales. 
% Previous studies \cite{du2023adaptive, huang2022ufpmp, li2020density, yang2019clustered} on aerial object detection have mainly relied on deep neural networks and coarse-to-fine architectures to enhance the detection performance. 
Previous studies \cite{du2023adaptive, huang2022ufpmp, li2020density, yang2019clustered, li2021gsdet, 10767172} on aerial object detection have predominantly employed deep neural networks and coarse-to-fine architectures to enhance detection performance. However, these methods, functioning as closed-set detectors, are inherently limited in their ability to recognize unseen categories.
Only a few works \cite{zang2024zero, li2024toward, li2024exploiting, pan2025locate} explore open-vocabulary detection for aerial images. 
DescReg \cite{zang2024zero} argues that aerial images exhibit a weak semantic-visual correlation and utilizes a triplet loss to preserve the visual similarity structure within the classification space, thereby enhancing knowledge transfer from base to novel classes.
CastDet \cite{li2024toward} employs a semi-supervised training paradigm within a complex multi-teacher framework. It leverages a pre-trained vision-language model as a teacher to generate high-quality pseudo-labels for unlabeled images containing novel classes, thereby enabling the detection of object categories beyond those present in the training dataset.
LAE-DINO \cite{pan2025locate} utilizes vision-language models to expand detectable categories and introduces a DINO-based detector that leverages dynamic vocabulary construction and visual-guided text prompt learning to explore class semantics, thereby enhancing OVD performance.
% These methods demonstrate the potential of the open-vocabulary approach in aerial object detection.

However, existing methods rely on complex architectures that overlook the efficiency demands in aerial detection and fail to fully exploit the benefits of image-text collaboration. 
In response, RT-OVAD builds upon a lighter framework and adopts a text-guided strategy to propagate class embeddings as clues throughout the entire network, helping the model focus on class-relevant object features and suppress extensive background interference.

\begin{figure*}[!t]
\centering
\includegraphics[width=.95\linewidth]{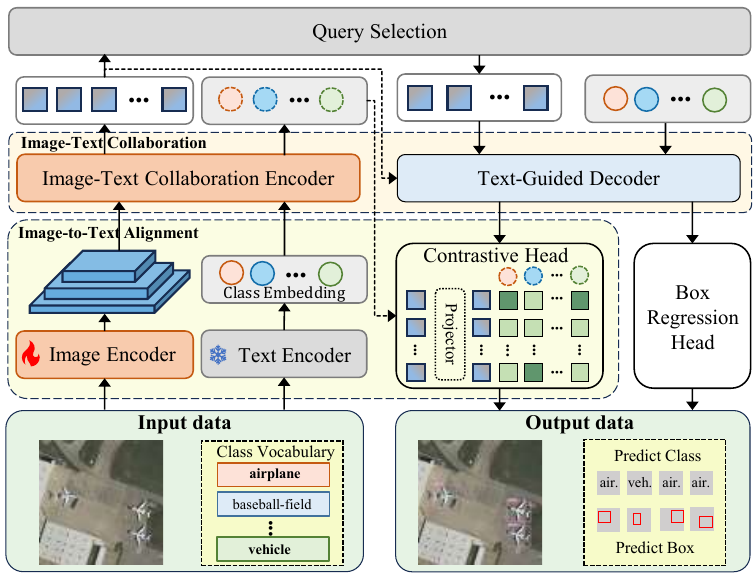}
\caption{
Overall architecture of RT-OVAD. The proposed RT-OVAD can be summarized into two main components: the Image-to-Text Alignment and the Image-Text collaboration Strategy.
(1) Image-to-Text Alignment connects visual objects with class embeddings rather than relying on class indices, which enables the model to detect classes beyond a predefined set. It involves a text encoder, a contrastive head, and a image-text alignment loss.
(2) Image-Text collaboration Strategy explores cross-modal interaction to enhance class-relevant feature extraction, suppress background interference and alleviate the impact of intra-class scale variations, forming a image-text collaboration encoder-decoder structure.
% Text-Guided Strategy explores cross-modal interaction to enhance class-relevant feature extraction, suppress background interference and alleviate the impact of intra-class scale variations, forming a text-guided encoder-decoder structure.
}
\label{overall}
\end{figure*}

\section{Method}
\subsection{Model Architecture}
The overall architecture of RT-OVAD is illustrated in Fig.~\ref{overall}. From the figure, we can see that the designs in the proposed RT-OVAD focus on two major components: the image-to-text alignment component and the image-text collaboration component. 
The former is proposed to overcome category limitations. We integrate class embeddings via a text encoder, constructing both a contrastive head and an image-to-text alignment loss to facilitate image-text alignment.
The latter is designed to enhance visual features, suppress background interference, and mitigate intra-class scale variations in aerial imagery. Specifically, we exploit image–text fusion between image features and class embeddings, forming an image–text collaborative encoder–decoder structure. This component consists of three key modules: Text-Guided Feature Enhancement (TG-FE), Visual-Guided Text Refinement (VG-TR), and Text-Guided Query Enhancement (TG-QE).

\subsection{Image-Text Alignment}

In this subsection, we focus on exploiting an effective image-text alignment approach to connect image and text features. Initially, we incorporate a text encoder to transform the class vocabulary into class embeddings. To further strengthen the connection between vision and language features, we then employ the corresponding image encoder to extract multi-scale image representations. Finally, we construct a contrastive head and introduce a image-to-text alignment loss, thereby aligning visual-semantic features in a comparable representation space.

\subsubsection{Text Encoder} 
To incorporate class semantics into the detector, we adopt the CLIP text encoder \cite{radford2021learning}. Since it contains very rich text features and is abundant for aerial image detection. The encoder accepts a class vocabulary \( C \in \mathbb{R}^n \) and produces class embeddings \( T = \text{TextEncoder}(C) \in \mathbb{R}^{n \times d} \), where \( n \) represents the number of class names and \( d \) denotes the dimensions of the class embeddings. 

\subsubsection{Image encoder}
Unlike natural images, the field of aerial object detection lacks sufficient data to support training an open-vocabulary detector from scratch. 
To enhance visual-semantic integration and bridge the connection between visual and text features, we also incorporate the CLIP image encoder \cite{radford2021learning}. Specifically, to ensure efficiency, we select the ResNet-50 version, modify it by removing the final AttnPool layer, and return the last three layers of the backbone as multi-scale feature outputs. Given an image \( I \in \mathbb{R}^{H \times W \times 3} \), the extracted image features \( \{F_3, F_4, F_5\} = \text{ImageEncoder}(I) \in \mathbb{R}^{H' \times W' \times C} \), where \( H', W', \) and \( C \) respectively denote the height, width, and channels of the feature maps.

\subsubsection{Image-to-Text Alignment Loss}
\textbf{Contrastive Head}.  
The contrastive head is designed to map visual-semantic features into a comparable representation space, producing similarity logit scores. 
Given the limited class vocabulary in aerial detection data and in order to preserve the generalization capability of the CLIP text encoder, we propose a image-to-text alignment strategy to bridge the domain gap between natural images and aerial images.
Specifically, we unidirectionally align the dimensions of the visual features to the corresponding class embeddings, keeping the class embeddings unaltered. Moreover, considering the demand for efficient detection, we adopt a simple linear layer to perform the image projection.

The visual-text similarity \( S(\mathbf{v}, \mathbf{t}) \) is computed by:
\begin{equation}
S(\mathbf{v}, \mathbf{t}) = \alpha \frac{\mathrm{proj}(\mathbf{v}) \cdot \mathbf{t}^T}{
\|\mathrm{proj}(\mathbf{v})\|_2 \cdot \|\mathbf{t}\|_2
} + \beta
\end{equation}
where \(\mathbf{v}\) is the query embedding, \(\mathbf{t}\) is the text embedding enhanced by VG-TR~(\ref{VG-TR}), and \(\mathrm{proj}(\cdot)\) denotes the linear layer used to align image features with class embeddings.
\(\alpha\) and \(\beta\) are learnable scale and offset parameters, respectively. 

This design allows us to fine-tune only a small number of learnable parameters, reducing optimization difficulty and the risk of overfitting, which is beneficial for overcoming the challenge of limited training samples in aerial imagery.

\noindent \textbf{Detection Loss}. The RT-OVAD loss consists of two components: a image-to-text alignment loss and a location loss. Specifically, given an image and a set of class vocabularies, RT-OVAD outputs \(N\) object predictions \(\{(b_i, s_i)\}_{i=1}^{N}\) for each decoder, where \(N\) denotes the number of queries. The corresponding ground-truth labels for these predictions are denoted as \(\{(b_k, u_k)\}_{k=1}^{K}\).

The image-to-text alignment loss is designed to align visual and textual features, replacing the traditional classification loss. Unlike the standard CLIP \cite{radford2021learning} pairwise image-text contrastive loss, we employ a pointwise contrastive loss that constrains the similarity score matrix between each query and the class embeddings, as obtained from the contrastive head. More specifically, considering the limited number of classes and the data constraints in aerial detection—and inspired by SigLIP \cite{zhai2023sigmoid}—we adopt sigmoid normalization instead of softmax. This conversion produces a pointwise contrastive loss and reduces dependencies among classes and batch sizes. The logits produced by the sigmoid function are then passed to the Varifocal loss \cite{zhang2021varifocalnet}, yielding the final image-to-text alignment loss \(L_{con}\). We formally define \(L_{con}\) as follows:
\begin{equation}
\label{eq:con_loss}
\small
\mathrm{L_{con}}(\theta, \mathbf{u}) = 
\begin{cases}
-\mathbf{u} \bigl(\mathbf{u} \log(\theta) + (1 - \mathbf{u})\log(1 - \theta)\bigr), & \text{if } \mathbf{u} > 0,\\
-\alpha \,\theta^{\gamma}\,\log\bigl(1 - \theta\bigr), & \text{if } \mathbf{u} = 0,
\end{cases}
\end{equation}
where \(\theta = \sigma(\mathbf{s})\), and \(\mathbf{s}\) denotes the similarity scores predicted by the contrastive head. For a positive sample, u for the ground-truth class is defined as the IoU between the predicted bounding box and its ground-truth bounding box, and 0 otherwise. For negative samples, \(u = 0\) for all classes.

Following previous works\cite{zhao2024detrs, zhang2022dino}, we adopt the GIoU loss \(L_{GIoU}\) and L1 loss \(L_{L1}\) for location regression. The training loss is formulated as: 
\begin{equation}
L_{det} = \lambda L_{con} + \mu L_{GIoU} + \nu L_{L1}
\end{equation}
where \(\lambda\), \(\mu\), and \(\nu\) represent the loss weights for the image-to-text alignment loss, GIOU loss, and L1 loss, respectively.

\subsection{Image-Text collaboration Strategy}
By incorporating the image-text alignment approach, we have already extended detection capabilities beyond predefined categories.
In this subsection, we further exploit image-text collaboration strategy to enhance image-text alignment and help the model focus on class-relevant features. Specifically, we propose Text-Guided Feature Enhancement (TG-FE) and Visual-Guided Text Refinement (VG-TR) within the encoder to simultaneously enhance visual features and refine textual embeddings. Additionally, we introduce a Text-Guided Query Enhancement (TG-QE) module within the decoder to enhance object queries, forming a complementary image–text collaboration encoder-decoder architecture.

\subsubsection{Image-Text Collaboration Encoder}
% As shown in~\ref{fig:TG-FE}, the image–text collaboration encoder comprises two modules: Text-Guided Feature Enhancement (TG-FE) and Visual-Guided Text Refinement (VG-TR).

\noindent \textbf{Text-Guided Feature Enhancement}. 
Due to the long imaging distance in aerial imagery, the images often contain small, low-resolution targets whose visual features are not prominent, making detection relatively challenging. Additionally, the broad coverage of aerial images means that the background may contain elements that strongly interfere with the detection process. To address these two issues, we propose using text guidance to enhance visual representations and design TG-FE to operate on the encoder part. 

\begin{figure}[!t]
\centering
\includegraphics[width=\linewidth]{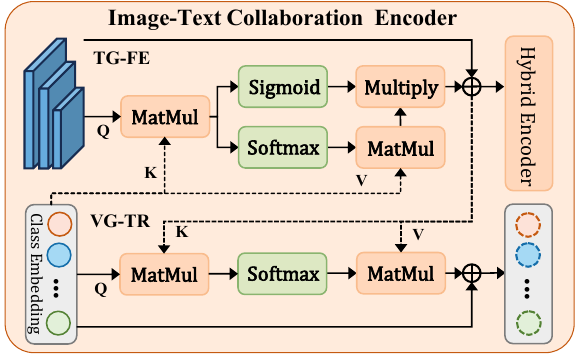}
\caption{The Image-Text collaboration Encoder in RT-OVAD. The TG-FE module enhances feature extraction by incorporating relevant class semantics into the visual features. The VG-TR module further refines the textual representations by fusing the enhanced visual features from TG-FE into the corresponding class embeddings.}
\label{fig:TG-FE}
\end{figure}

As illustrated in Fig. \ref{fig:TG-FE}, the core idea is based on cross attention. Specifically, given multi-scale image features \( F_{i} \in \mathbb{R}^{H \times W \times C} \) from the image encoder and class embeddings \( T \in \mathbb{R}^{N \times D} \), we employ cross-attention to fuse these modalities. To suppress the excessive injection of class embeddings into background features, we compute the maximum similarity between each image feature and the class embeddings, then apply a sigmoid function to interpret this as the probability that the image feature belongs to the foreground. We then multiply the text feature representation \( T_{j}^{(v)} \) by this probability. The image features are updated as follows:
\begin{equation}
\label{eq:attn}
\text{Attn} = \frac{F_{i}^{(q)} T_{j}^{(k)T}}{\sqrt{d}},
\end{equation}
\begin{equation}
\label{eq:update}
F_{i}' = F_{i} + \phi(\text{Attn}) \cdot T_{j}^{(v)} \cdot \sigma\Bigl(\max_{j=1, \ldots, n}(\text{Attn})\Bigr),
\end{equation}
where \(F_{i}^{(q)}\) denotes the query projection of the \(i\)-th layer image feature in the multi-scale features, \(T_{j}^{(k)}\) and \(T_{j}^{(v)}\) represent the key and value projections of the text embeddings, \(\phi(\cdot)\) denotes the softmax function, and \(\sigma(\cdot)\) denotes the sigmoid function. Additionally, to reduce the parameter count and ensure a lightweight design, features of different scales share the same set of TG-FE parameters.

The softmax branch integrates text information into the image features, which can improve issues related to inconspicuous target visual features. Meanwhile, the sigmoid branch, guided by text information, functions like a gating mechanism, suppressing irrelevant background interference during detection. Fig. \ref{fig:TG-FE-impact} visually demonstrates the impact of TG-FE on the features.

\begin{figure}[!t]
\centering
\includegraphics[width=\linewidth]{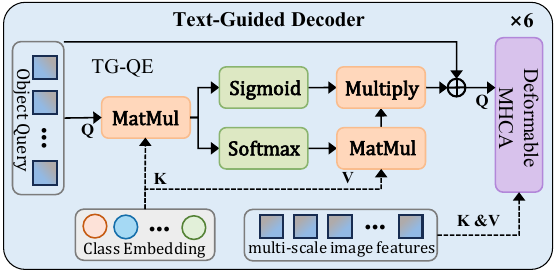} 
\caption{The Text-Guided Decoder in RT-OVAD. By incorporating class information into object queries, TG-QE enables more focused retrieval of class-relevant features.}
\label{fig:TG-QE}
\end{figure}

\noindent \textbf{Visual-Guided Text Refinement}.
\label{VG-TR}
% Due to varying viewpoints and shooting altitudes, aerial images often exhibit significant object scale variations. As a result, objects belonging to the same category may present drastically different visual appearances across scenes, making it challenging to align them with unified class embeddings. A straightforward solution is to enhance class embeddings with corresponding visual features. However, this approach may inevitably introduce irrelevant visual information—such as background interference—into the class representations.
Due to varying viewpoints and shooting altitudes, aerial images often exhibit significant object scale variations. As a result, objects belonging to the same category may present drastically different visual appearances across scenes, making it challenging to align them with unified class embeddings. A straightforward solution is to enhance class embeddings with corresponding visual features. However, this approach may inevitably introduce irrelevant visual information—such as background interference—into the class representations.

% To address this, we introduce a Visual-Guided Text Refinement (VG-TR) module following TG-FE. Specifically, VG-TR employs a cross-attention mechanism to fuse the enhanced visual features produced by TG-FE with the textual embeddings. 
To address this, we introduce a Visual-Guided Text Refinement (VG-TR) module following TG-FE. Specifically, VG-TR employs a cross-attention mechanism to fuse the enhanced visual features produced by TG-FE with the textual embeddings. Formally, given the refined visual features \( F'_{i} \in \mathbb{R}^{H \times W \times C} \) generated by the TG-FE module, and textual embeddings \( T_j \in \mathbb{R}^{N \times D} \), the VG-TR module updates textual embeddings as follows:
\begin{equation}
\label{eq:vgtr_attn}
\text{Attn} = \frac{T_j^{(q)} {F'_{i}}^{(k)T}}{\sqrt{d}},
\end{equation}
\begin{equation}
\label{eq:vgtr_update}
T_j' = T_j + \phi(\text{Attn}) \cdot {F'_{i}}^{(v)},
\end{equation}
where \(T_j^{(q)}\) denotes the query projection of the textual embeddings, \({F'_{i}}^{(k)}\) and \({F'_{i}}^{(v)}\) represent the key and value projections of the refined visual features from TG-FE, respectively, and \(\phi(\cdot)\) denotes the softmax function.

As a result, the generated text embeddings incorporate class-relevant instance features and contextual scene information, enabling the model to more effectively capture intra-class variations in appearance and scale. Furthermore, since the visual features used in VG-TR have already been refined by TG-FE—which explicitly suppresses background interference—the risk of contaminating class embeddings with irrelevant visual cues is substantially reduced. Ultimately, these refined and context-aware textual embeddings are utilized to compute visual-text similarity within the contrastive head.

% As a result, the generated text embeddings incorporate class-relevant instance features and contextual scene information, enabling the model to more effectively capture intra-class variations in appearance and scale. Furthermore, since the visual features used in VG-TR have already been refined by TG-FE—which explicitly suppresses background interference—the risk of contaminating class embeddings with irrelevant visual cues is substantially reduced. Ultimately, these refined and context-aware textual embeddings are utilized to compute visual-text similarity within the contrastive head.

\subsubsection{Text-Guided Query Enhancement}
Similarly to TG-FE, we propose TG-QE, which operates on the decoder part. As shown in Fig.~\ref{fig:TG-QE}, TG-QE enriches object queries with class embeddings as prior knowledge so that, when these queries interact with multi-scale image features in the decoder, they can more effectively focus on class-relevant features and thus improve detection performance.

However, since the number of object queries is generally larger than the number of actual objects, most queries end up being negative samples. To avoid interference with label assignment, TG-QE adaptively suppress the injection of excessive class embeddings into these negative queries. Specifically, the object queries are updated as follows:
\begin{equation}
\label{eq:attn1}
\text{Attn} = \frac{Q_{i}^{(q)} T_{j}^{(k)T}}{\sqrt{d}},
\end{equation}
\begin{equation}
\label{eq:update1}
Q_{i}' = Q_{i} + \phi(\text{Attn}) \cdot T_{j}^{(v)} \cdot \sigma\Bigl(\max_{j=1, \ldots, n}(\text{Attn})\Bigr),
\end{equation}
where \(Q_{i}^{(q)}\) denotes the query projection of the \(i\)-th layer object query features in the multi-scale features, \(T_{j}^{(k)}\) and \(T_{j}^{(v)}\) represent the key and value projections of the text embeddings, \(\phi(\cdot)\) denotes the softmax function, and \(\sigma(\cdot)\) denotes the sigmoid function.

\section{Experiments}
In this section, we aim to demonstrate that the proposed RT-OVAD has the ability to break category limitations and explicitly demonstrate the effectiveness of our proposed components. To comprehensively evaluate our method, we conduct experiments on three benchmark tasks: open-vocabulary aerial object detection, zero-shot aerial detection, and traditional closed-set object detection.

\subsection{Datasets and Experiment Setup} \label{sec:datasets}
\noindent \textbf{Datasets}.
For the evaluation of open-vocabulary aerial object detection, we follow the benchmark protocol established in \cite{pan2025locate}, where the model is first pretrained on LAE-1M and then evaluated on three benchmark datasets: DIOR \cite{li2020object}, DOTA-v2.0 \cite{xia2018dota}, and LAE-80C \cite{pan2025locate}.

For the evaluation of zero-shot aerial object detection, we use three widely-adopted benchmark datasets: xView \cite{lam2018xview}, DIOR \cite{li2020object}, and DOTA \cite{xia2018dota}. To ensure fairness and validity in our comparative experiments, we propose two distinct protocols for splitting the datasets into base and novel classes. In the first protocol, we follow the setup proposed in DescReg \cite{zang2024zero}. However, we observe that some novel classes in DIOR overlap with the base classes in DOTA, and that DOTA-v1.0 lacks sufficient annotations for small instances. To avoid class leakage and mitigate annotation inconsistencies, in the second protocol we adopt DOTAv1.5 and adjust the category splits in DOTA accordingly. Full details of the base/novel category splits used for zero-shot evaluation are provided in the appendix.

For conventional closed-set aerial object detection, RT-OVAD is further evaluated on two standard UAV benchmarks: VisDrone~\cite{du2019visdrone} and UAVDT~\cite{du2018unmanned}.

\noindent \textbf{Evaluation Metrics}. Following \cite{zang2024zero, pan2025locate}, we adopt standard object detection metrics for evaluation, including mean Average Precision (mAP), AP${50}$, Recall, Harmonic Mean (HM), and Frames Per Second (FPS).
The mAP measures the average precision across multiple IoU thresholds ranging from 0.5 to 0.95 (in steps of 0.05). AP${50}$ and Recall are computed using an Intersection over Union (IoU) threshold of 0.5. The HM metric is used to reflect the overall performance across base and novel classes, and is computed as the harmonic mean of their respective scores. Additionally, for zero-shot detection tasks, we report results under two evaluation settings: Zero-shot Detection (ZSD), which evaluates performance on novel classes only, and Generalized Zero-shot Detection (GZSD), which assesses performance on both base and novel classes simultaneously.

\noindent \textbf{Implementation Details}
RT-OVAD is implemented based on RT-DETR \cite{zhao2024detrs} with a ResNet50 backbone \cite{he2016deep}. By incorporating an additional text encoder and contrastive loss, we form an end-to-end detection framework. All experiments were conducted on four 4090 GPUs, each with a batch size of 2. We set the number of object queries to 500 to accommodate the higher number of objects typically found in aerial images compared to natural scenes. Unless otherwise specified, the frames per second (FPS) are reported for a single 4090 GPU without the use of any additional acceleration techniques. Due to the space limitation, further details on the implementation are provided in the supplementary material.

\begin{table*}[t]
\caption{Comparison with state-of-the-art open-vocabulary detectors on DIOR, DOTAv2.0, and LAE-80C benchmarks. 
All models are pre-trained on LAE-1M. FPS is tested on a 4090 GPU without additional acceleration. The \textbf{best} result within each split is shown in bold, and the \underline{second-best} is underlined.}
\label{tab:OVAD}
\centering
\small
\setlength{\tabcolsep}{1.5mm}
\begin{tabular}{@{}ccccccccc@{}}
\toprule
\multirow{2}{*}{\textbf{Method}} & 
\multirow{2}{*}{\textbf{Source}} & 
\multirow{2}{*}{\textbf{Pre-training Data}} & 
\multirow{2}{*}{\textbf{Resolution}} & 
\multirow{2}{*}{\textbf{Params}} & 
\multirow{2}{*}{\textbf{FPS}} & 
\multicolumn{1}{c}{\textbf{DIOR}} & 
\multicolumn{1}{c}{\textbf{DOTAv2.0}} & 
\multicolumn{1}{c}{\textbf{LAE-80C}} \\
\cmidrule(lr){7-9}
& & & & & & AP$_{50}$ & mAP & mAP \\
\midrule
YOLO-World~\cite{cheng2024yolo}         & CVPR 2024 & LAE-1M & 800$\times$800  & 110M & 35  & 78.0 & 33.5 & 13.6 \\
Grounding DINO~\cite{liu2024grounding}  & ECCV 2024 & LAE-1M & 1333$\times$800 & 172M & 11  & 83.6 & 46.0 & 17.7 \\
LAE-DINO~\cite{pan2025locate}           & AAAI 2025 & LAE-1M & 1333$\times$800 & 180M & 10  & \underline{85.5} & \underline{46.8} & \underline{20.2} \\
\rowcolor{gray!15}
\textbf{RT-OVAD (Ours)}                 & —         & LAE-1M & 800$\times$800  & 54M  & 34  & \textbf{87.7} \textcolor{green!60!black}{\small(+2.2)} & \textbf{53.8} \textcolor{green!60!black}{\small(+7.0)} & \textbf{23.7} \textcolor{green!60!black}{\small(+3.5)} \\
\bottomrule
\end{tabular}
\end{table*}

\begin{table*}[t]
\caption{Comparison with the state-of-the-art under GZSD and ZSD on the xView dataset, where * denotes that the dataset split redefines the base and novel classes to avoid leakage of novel classes. The \textbf{best} result within each split is shown in bold, and the \underline{second-best} is underlined.}
\label{xView}
\centering
\setlength{\tabcolsep}{2.mm}
\begin{tabular}{@{}cccc ccc ccc cc@{}}
\toprule
\multirow{3}{*}{\textbf{Method}} &
\multirow{3}{*}{\textbf{Source}} &
\multirow{3}{*}{\textbf{Backbone}} &
\multirow{3}{*}{\textbf{Resolution}} &
\multicolumn{6}{c}{\textbf{GZSD}} &
\multicolumn{2}{c}{\textbf{ZSD}} \\
\cmidrule(lr){5-10}\cmidrule(lr){11-12}
& & & &
\multicolumn{3}{c}{\textbf{AP$_{50}$}} &
\multicolumn{3}{c}{\textbf{Recall}} &
\multirow{2}{*}{\textbf{AP$_{50}$}} &
\multirow{2}{*}{\textbf{Recall}} \\
\cmidrule(lr){5-7}\cmidrule(lr){8-10}
& & & &
Base & Novel & HM & Base & Novel & HM & & \\
\midrule
RRFS\cite{huang2022robust}            & CVPR22 & ResNet\textendash101 & 800$\times$800 & 10.2 & 1.6 & 2.7 & 19.1 & 5.8 & 8.9 & 2.2 & 14.3 \\
ConstrastZSD\cite{yan2022semantics}   & TPAMI22& ResNet\textendash101 & 800$\times$800 & 16.8 & 2.9 & 5.0 & 27.6 & 13.9 & 18.5 & 4.1 & 27.1 \\
DescReg\cite{zang2024zero}            & AAAI24 & ResNet\textendash101 & 800$\times$800 & 17.1 & \textbf{5.8} & \textbf{8.7} & 28.0 & 12.8 & 17.6 & 8.3 & 43.0 \\
YOLO-World\cite{cheng2024yolo}        & CVPR24 & YOLOv8-L   & 800$\times$800 & 18.5 & 3.3 & 5.6 & 41.4 & 15.2 & 22.2 & 7.9 & 37.1 \\
Grounding DINO\cite{liu2024grounding} & ECCV24 & Swin-T     & 1333$\times$800 & 20.3 & 4.8 & 7.7 & \underline{47.0} & 22.1 & 30.0 & \underline{8.7} & \underline{58.6} \\
LAE-DINO\cite{pan2025locate}          & AAAI25 & Swin-T     & 1333$\times$800 & \underline{20.5} & 4.7 & 7.6 & 46.2 & \underline{31.2} & \underline{37.3} & 7.2 & 58.3 \\
\rowcolor{gray!20}
\textbf{RT-OVAD (Ours)}               & --     & ResNet\textendash50  & 800$\times$800 & \textbf{21.1} & \underline{5.3} & \underline{8.5} & \textbf{58.9} & \textbf{34.9} & \textbf{43.8} & \textbf{9.8} & \textbf{60.1} \\
\midrule
YOLO-World*                           & CVPR24 & YOLOv8-L   & 800$\times$800 & \underline{21.7} & 3.2 & 5.6 & 46.6 & 16.5 & 24.4 & 8.8 & 41.7 \\
Grounding DINO*                       & ECCV24 & Swin-T     & 1333$\times$800 & 20.9 & \underline{6.2} & \underline{9.5} & 47.8 & \underline{31.1} & \underline{37.7} & \underline{11.1} & \textbf{63.5} \\
LAE-DINO*                             & AAAI25 & Swin-T     & 1333$\times$800 & 21.4 & 4.6 & 7.6 & \underline{50.1} & 26.0 & 34.0 & 8.9 & 58.8 \\
\rowcolor{gray!20}
\textbf{RT-OVAD (Ours)*}              & --     & ResNet\textendash50  & 800$\times$800 & \textbf{22.4} & \textbf{7.0} & \textbf{10.6} & \textbf{60.1} & \textbf{36.7} & \textbf{45.5} & \textbf{12.3} & \underline{62.5} \\
\bottomrule
\end{tabular}
\end{table*}

\begin{table*}[t]
\caption{Comparison with the state-of-the-art under GZSD and ZSD on the DIOR and DOTA datasets.  
Two evaluation protocols are reported: the upper block adopts the standard base/novel split, while the lower block (marked with~*) follows a re-defined split to avoid novel-class leakage.  
The \textbf{best} value is shown in bold and the \underline{second-best} is underlined.}
\label{DIOR and DOTA}
\centering
\setlength{\tabcolsep}{1.5mm}
\begin{tabular}{@{}ccccccccccccccccc@{}}
\toprule
\multirow{4}{*}{\textbf{Method}} &
\multicolumn{8}{c}{\textbf{DIOR}} &
\multicolumn{8}{c}{\textbf{DOTA}} \\ \cmidrule(lr){2-9}\cmidrule(lr){10-17}
& \multicolumn{6}{c}{\textbf{GZSD}} & \multicolumn{2}{c}{\textbf{ZSD}} 
& \multicolumn{6}{c}{\textbf{GZSD}} & \multicolumn{2}{c}{\textbf{ZSD}} \\ 
\cmidrule(lr){2-7}\cmidrule(lr){8-9}\cmidrule(lr){10-15}\cmidrule(lr){16-17}
& \multicolumn{3}{c}{\textbf{AP$_{50}$}} & \multicolumn{3}{c}{\textbf{Recall}} & 
\multirow{2}{*}{\textbf{AP$_{50}$}} & \multirow{2}{*}{\textbf{Recall}} 
& \multicolumn{3}{c}{\textbf{AP$_{50}$}} & \multicolumn{3}{c}{\textbf{Recall}} & 
\multirow{2}{*}{\textbf{AP$_{50}$}} & \multirow{2}{*}{\textbf{Recall}} \\ 
\cmidrule(lr){2-4}\cmidrule(lr){5-7}\cmidrule(lr){10-12}\cmidrule(lr){13-15}
& Base & Novel & HM & Base & Novel & HM & & & Base & Novel & HM & Base & Novel & HM & & \\
\midrule
RRFS                         & 41.9 & 2.8 & 5.2 & 60.0 & 19.9 & 29.9 & 9.7 & 19.8 & 47.1 & 2.2 & 4.2 & 71.4 & 14.2 & 23.7 & 2.9 & 14.4 \\
ConstrastZSD                 & 51.4 & 3.9 & 7.2 & 69.2 & 25.9 & 37.7 & 8.7 & 22.3 & 41.6 & 2.8 & 5.2 & 69.1 & 12.2 & 20.7 & 6.0 & 25.4 \\
DescReg                      & 68.7 & 7.9 & 14.2 & 82.0 & 34.3 & 48.4 & 15.2 & 34.6 & 68.7 & 4.7 & 8.8 & 83.8 & 29.9 & 44.0 & 8.5 & 34.4 \\
YOLO-World-L                 & \underline{80.2} & 17.3 & 28.5 & 91.1 & 38.2 & 53.8 & 24.8 & 37.8 & 74.4 & 22.7 & 34.8 & 87.8 & 42.2 & 57.0 & 24.3 & 35.7 \\
Grounding DINO               & \textbf{80.7} & \underline{29.4} & \underline{43.1} & \underline{94.1} & \textbf{63.8} & \textbf{76.0} & \underline{37.7} & \textbf{84.4} & \textbf{77.5} & \underline{23.5} & \underline{36.1} & \underline{92.8} & \underline{65.4} & \underline{76.7} & \underline{28.1} & \underline{76.4} \\
LAE-DINO                     & 79.4 & \textbf{32.2} & \textbf{45.9} & 93.5 & \underline{63.5} & \underline{75.7} & \textbf{38.2} & \underline{74.9} & \underline{76.8} & 22.9 & 35.3 & 91.5 & 56.4 & 69.7 & 24.2 & 73.1 \\
\rowcolor{gray!20}
RT-OVAD                      & 80.0 & 23.2 & 36.0 & \textbf{94.6} & 62.5 & 75.2 & 37.5 & 74.2 & 76.3 & \textbf{24.0} & \textbf{36.5} & \textbf{93.8} & \textbf{74.1} & \textbf{82.8} & \textbf{32.4} & \textbf{80.3} \\
\midrule
YOLO-World-L*                & \underline{78.5} & 3.1 & 5.9 & 91.5 & 18.6 & 30.9 & 6.7 & 20.6 & 71.5 & 1.7 & 3.3 & 80.9 & 27.1 & 40.6 & 4.2 & 19.5 \\
Grounding DINO*              & 78.4 & 6.0 & 11.2 & \underline{93.9} & \underline{65.3} & \underline{77.0} & 9.8 & \underline{81.0} & 71.2 & 2.8 & 5.4 & \textbf{86.4} & 47.8 & 61.6 & 7.5 & 75.0 \\
LAE-DINO*                    & \textbf{79.0} & \underline{6.6} & \underline{12.0} & 93.8 & \textbf{66.5} & \textbf{77.8} & \underline{10.4} & 80.7 & \underline{71.9} & \textbf{4.8} & \textbf{9.0} & \textbf{86.4} & \underline{51.7} & \underline{64.6} & \underline{8.0} & \textbf{76.7} \\
\rowcolor{gray!20}
RT-OVAD*                     & \underline{78.5} & \textbf{7.6} & \textbf{13.9} & \textbf{94.2} & 64.9 & 76.8 & \textbf{14.1} & \textbf{81.2} & \textbf{72.2} & \underline{2.9} & \underline{5.6} & \underline{85.9} & \textbf{66.6} & \textbf{74.9} & \textbf{9.6} & \underline{76.6} \\
\bottomrule
\end{tabular}
\end{table*}

\subsection{Performance on Open-Vocabulary Aerial Detection}
As shown in Table~\ref{tab:OVAD}, we evaluate the proposed RT-OVAD on the open-vocabulary aerial detection benchmark established by~\cite{pan2025locate}, achieving state-of-the-art (SOTA) performance across three benchmark datasets. Specifically, we compare our method with three mainstream open-vocabulary detectors designed for natural images (GLIP, Grounding DINO, and YOLO-World), as well as the recent aerial-specific detector LAE-DINO. The experimental results demonstrate that, despite using a lower-resolution input of $800\times800$, our RT-OVAD still significantly outperforms previous SOTA methods, surpassing them by \textbf{+2.2}, \textbf{+7.0}, and \textbf{+3.5} in terms of AP$_{50}$ on the DIOR dataset, and mAP on DOTAv2.0 and LAE-80C datasets, respectively.

\noindent \textbf{Model Size and Latency}
Table~\ref{tab:OVAD} also compares model complexity and inference speed. In particularly, compared to other methods, RT-OVAD uses fewer parameters while achieving the highest AP$_{50}$/mAP performance on three benchmark datasets. RT-OVAD also achieves an inference speed of 34 FPS on an NVIDIA RTX 4090 GPU, notably faster than Grounding DINO (11 FPS), LAE-DINO (10 FPS), and comparable to YOLO-World (35 FPS), and meeting the real-time detection requirements for various applications. 
These results highlight that RT-OVAD successfully inherits the efficiency of its base architecture, making it particularly well-suited for open scenarios and aerial detection. 

\subsection{Performance on Zero-shot Aerial Detection}
As mentioned in Section~\ref{sec:datasets}, we adopt two protocols for splitting the base and novel classes. The second protocol is designed to avoid class leakage, and its results are marked with an asterisk (*) in the subsequent tables. For a comprehensive evaluation, we compare our approach not only with state-of-the-art aerial zero-shot detectors but also with the effective open-vocabulary detection method YOLO-World~\cite{cheng2024yolo}, Grounding DINO~\cite{liu2024grounding} and LAE-DINO~\cite{pan2025locate}.

\begin{table}[t]
\caption{Comparison with the state-of-the-art on the VisDrone validation set.
“800–1200” indicates multi-scale inputs. The \textbf{best} value is shown in bold and the \underline{second-best} is underlined.}
\label{visdrone}
\centering
\small
\setlength{\tabcolsep}{4.2pt}
\begin{tabular}{@{}cccc@{}}
\toprule
\textbf{Method} & \textbf{Backbone} & \textbf{Resolution} & \textbf{AP$_{50}$} \\
\midrule
DMNet\cite{li2020density}          & ResNet-50 & 1000$\times$600  & 47.6 \\
QueryDet\cite{yang2022querydet}    & ResNet-50 & 2400$\times$2400 & 48.1 \\
ClusDet\cite{yang2019clustered}    & ResNet-50 & 1000$\times$600  & 50.6 \\
CZ Det\cite{meethal2023cascaded}   & ResNet-50 & 800–1200         & 58.3 \\
RT-DETR\cite{zhao2024detrs}        & ResNet-50 & 800$\times$800   & 58.9 \\
UFPMP-Det\cite{huang2022ufpmp}     & ResNet-50 & 1333$\times$800  & \underline{62.4} \\
\midrule
GFL V1 (CEASC)\cite{du2023adaptive} & ResNet-18 & 1333$\times$800  & 50.7 \\
YOLO-World\cite{cheng2024yolo}      & YOLOv8-L  & 800$\times$800   & 45.0 \\
YOLO-World\cite{cheng2024yolo}      & YOLOv8-L  & 1280$\times$1280 & 55.3 \\
LAE-DINO\cite{pan2025locate}      & Swin-T  & 1333$\times$800   & 56.4 \\
Grounding DINO\cite{liu2024grounding}  & Swin-T  & 1333$\times$800   & 58.1 \\
\midrule
\rowcolor{gray!20}
RT-OVAD                             & ResNet-50 & 800$\times$800   & 60.1 \\
\rowcolor{gray!20}
RT-OVAD                             & ResNet-50 & 1280$\times$1280 & \textbf{64.6} \\
\bottomrule
\end{tabular}
\end{table}

\noindent \textbf{Evaluation on xView.} 
In Tab.~\ref{xView}, we compare the performance of RT-OVAD with state-of-the-art methods on the \textbf{xView} dataset. 
% Under the ZSD setting, RT-OVAD achieves the highest AP${50}$ 9.8\% and Recall 60.1\% in the first protocol, outperforming Grounding DINO and LAE-DINO by +1.1\% and +2.6\% in AP${50}$, respectively, and by +1.5\% and +1.8\% in Recall. 
Under the ZSD setting, RT-OVAD achieves the highest AP${50}$ of 9.8\% and Recall of 60.1\% in the first protocol, outperforming Grounding DINO (+1.1\% AP${50}$, +1.5\% Recall), LAE-DINO (+2.6\% AP${50}$, +1.8\% Recall), and YOLO-World (+1.9\% AP${50}$, +23.0\% Recall). In the second protocol, RT-OVAD achieves an AP$_{50}$ of 12.3\%, surpassing all other compared methods.

In the GZSD setting, RT-OVAD consistently outperforms all compared methods in Recall across both base and novel classes. It achieves a novel-class AP${50}$ of 7.0\% and Recall of 36.7\% in the second protocol, achieving the highest HM in both AP and Recall.

% \begin{table}[t]
% \caption{Comparison of various detection methods on the DIOR dataset. Speed is tested on a 4090 GPU with no additional acceleration. Results of DescReg are taken from its original paper. }
% \label{FPS}
% \centering
% % \setlength{\tabcolsep}{1.0mm} 
% \begin{tabular}{c|c|c|c}
% \hline
% {\textbf{Method}} & {\textbf{Params}} & {\textbf{ZSD}} & {\textbf{FPS}$_{bs=1}$} \\
% % \cline{2-3}
%  % & {\textbf{GZSD}} & {\textbf{ZSD}} & \\
% \hline
% DescReg(Faster-RCNN) & $\geq$ 61M & 15.2 & 11 \\
% % Yolo-World-M & 92M & 23.4 &  - \\
% Yolo-World-L & 110M & 24.8 &  35 \\
% % Grounding DINO & - & - & - \\
% \hline
% RT-OVAD(ResNet-50) & \textbf{49M} & \textbf{37.2} & \textbf{36} \\
% \hline
% \end{tabular}
% \end{table}

\noindent \textbf{Evaluation on DIOR and DOTA.} 
% In Tab.~\ref{DIOR and DOTA}, we present the performance of RT-OVAD on the \textbf{DIOR} and \textbf{DOTA} datasets. In the ZSD setting, RT-OVAD demonstrates remarkable improvements in both Recall and mAP across both splits, the recall metric surpassing previous best results by 42.0\%/55.3\% on DIOR and 45.6\%/59.3\% on DOTA for the first/second split, respectively. Likewise, under the GZSD setting, RT-OVAD achieves a higher HM of Recall than all compared approaches under both protocol splits, highlighting its strong capability for detecting novel categories. Extensive experiments across three benchmark datasets confirm that RT-OVAD achieves notable performance gains in both ZSD and GZSD settings, especially in terms of Recall. These results underscore that RT-OVAD offers more robust open-vocabulary detection capabilities in aerial detection. 
We further evaluate RT-OVAD on the \textbf{DIOR} and \textbf{DOTA} datasets under both ZSD and GZSD settings, as shown in Tab.~\ref{DIOR and DOTA}. On DIOR, RT-OVAD* achieves the highest ZSD AP$_{50}$ of 14.1\%, surpassing LAE-DINO* and Grounding DINO* by 3.7\% and 4.3\%, respectively. In terms of Recall, RT-OVAD* reaches 81.2\%, outperforming all compared methods and demonstrating strong generalization to novel classes. 
In the GZSD setting, RT-OVAD also achieves the highest harmonic mean (HM) of AP${50}$ at 13.9\%, representing a +1.9\% improvement over LAE-DINO*. On the DOTA dataset, RT-OVAD* achieves the best ZSD AP${50}$ of 9.6\% and competitive Recall. While RT-OVAD exhibits slightly lower performance than LAE-DINO in a few GZSD metrics, it demonstrates clear advantages in model efficiency: as shown in Tab.~\ref{tab:OVAD}, RT-OVAD significantly reduces parameter count and inference latency compared to both Grounding DINO and LAE-DINO. These results underscore that RT-OVAD offers more robust open-vocabulary detection capabilities in aerial detection. 

\subsection{Performance on Traditional Aerial Detection}
\noindent \textbf{Performance on Visdrone dataset.} 
Although RT-OVAD is designed for open vocabulary scenarios, it can also be applied to traditional aerial object detection. To this end, we conducted experiments on Visdrone to assess its performance on traditional aerial object detection, and list the results in Tab.~\ref{visdrone}. For a fair comparison with previous work, we tested using different input resolutions. At an input size of $800\times800$, RT-OVAD achieved an mAP of 60.1\%, an improvement of 1.2\% compared to RT-DETR. Increasing the resolution to $1280\times1280$ boosted the mAP by 2.2\% beyond the previously best-reported result, achieving state-of-the-art performance. 

\noindent \textbf{Performance on UAVDT dataset.} 
UAVDT \cite{du2018unmanned} is also a benchmark for evaluation in aerial object detection. The dataset comprises 23,258 training images and 15,069 testing images with a resolution of \(1024 \times 540\), covering 3 classes.
Besides Visdrone, to further demonstrate the performance of RT-OVAD on traditional aerial detection tasks, we evaluate our method on the UAVDT~\cite{du2018unmanned} dataset. As shown in Table~\ref{uavdt}, RT-OVAD achieves the highest mAP performance, surpassing the best-compared method by 1.0\% in mAP. In summary, our proposed RT-OVAD not only adapts to open-vocabulary aerial object detection scenarios but also retains excellent performance in closed settings. 

\subsection{Ablation Study}
We conduct ablation experiments on the DIOR dataset to assess the contribution of each proposed component, as shown in Tab.~\ref{ablation_study_zsd}. The baseline incorporates a text encoder and a vanilla contrastive loss for image-text alignment, which enables zero-shot detection of novel classes.

\begin{table}[!t]
\caption{Comparison with the state-of-the-art on the UAVDT dataset. The \textbf{best} value is shown in bold and the \underline{second-best} is underlined.}
\label{uavdt}
\centering
\small
\setlength{\tabcolsep}{4.2pt}
\begin{tabular}{@{}cccc@{}}
\toprule
\textbf{Method} & \textbf{Backbone} & \textbf{Resolution} & \textbf{AP$_{50}$} \\
\midrule
DMNet\cite{li2020density}        & ResNet-50 & 1000$\times$600  & 24.6 \\
ClusDet\cite{yang2019clustered}  & ResNet-50 & 1000$\times$600  & 26.5 \\
GLSAN\cite{deng2020global}       & ResNet-50 & 1000$\times$600  & 28.1 \\
ARMNet\cite{wei2020amrnet}       & ResNet-50 & 1500$\times$800  & 30.4 \\
UFPMP-Det\cite{huang2022ufpmp}   & ResNet-50 & 1000$\times$600  & 38.7 \\
RT-DETR\cite{zhao2024detrs}      & ResNet-50 & 800$\times$800   & \underline{39.5} \\
\midrule
GFL V1 (CEASC)\cite{du2023adaptive}   & ResNet-18 & 1024$\times$540  & 30.9 \\
YOLO-World\cite{cheng2024yolo}        & YOLOv8-L  & 800$\times$800   & 35.8 \\
Grounding DINO\cite{liu2024grounding} & Swin-T    & 1333$\times$800  & 36.0 \\
LAE-DINO\cite{pan2025locate}          & Swin-T    & 1333$\times$800  & 36.5 \\
\midrule
\rowcolor{gray!20}
RT-OVAD                             & ResNet-50 & 800$\times$800   & \textbf{40.5} \\
\bottomrule
\end{tabular}
\end{table}

\noindent \textbf{Ablation on Image-to-Text Alignment.} To facilitate image-text contrastive learning, we adopt CLIP-based encoders for both images and text. This improves the GZSD mAP (HM) by 3.1\% and the ZSD mAP by 3.6\%. Given the limited aerial detection data, we propose the image-to-text alignment loss to alleviate the challenge of image-text alignment and preserve the text encoder’s generalization capability. This further boosts the GZSD mAP (HM) and ZSD mAP by 3.2\% and 2.8\%, respectively. These results suggest that our proposed method effectively enhances the capability to detect novel classes in the aerial detection.

\noindent \textbf{Ablation on Image-Text Collaboration Strategy.} 
As illustrated in Tab.~\ref{ablation_study_zsd}, introducing TG-FE or TG-QE individually yields notable performance improvements across both experimental settings. When both modules are employed together to form a text-guided encoder-decoder architecture, the GZSD mAP (HM) and ZSD mAP achieve 13.1\% and 14.3\%, respectively. 
Furthermore, when the VG-TR is introduced on top of TG-FE and TG-QE, the model obtains the highest GZSD HM of 13.9\%, validating that enriching text embeddings with context-aware visual cues further boosts image-text alignment and improves generalization to novel classes.
These results highlight the complementary benefits of the TG-FE, TG-QE, and VG-TR modules, and confirm the overall effectiveness of the proposed image-text collaboration design.

\begin{table}[t]
\caption{Ablation study of each component on the DIOR dataset.
C-WL: loading CLIP encoder weights; I2T Loss: image-to-text alignment loss.}
\label{ablation_study_zsd}
\centering
\setlength{\tabcolsep}{4.2pt}
\small
\begin{tabular}{@{}ccccccc@{}}
\toprule
\multicolumn{2}{c}{\textbf{I2T Align}} &
\multicolumn{3}{c}{\textbf{I-T Collaboration Strategy}} &
\multicolumn{1}{c}{\textbf{GZSD}} &
\multicolumn{1}{c}{\textbf{ZSD}} \\
\cmidrule(lr){1-2}\cmidrule(lr){3-5}\cmidrule(lr){6-7}
C-WL & I2T Loss & TG-FE & TG-QE & VG-TR & HM & AP$_{50}$ \\
\midrule
          &           &           &           &           &  2.5 &  3.4 \\
\checkmark&           &           &           &           &  5.6 &  7.0 \\
\checkmark& \checkmark&           &           &           &  8.8 &  9.8 \\
\checkmark& \checkmark& \checkmark&           &           & 10.4 & 12.3 \\
\checkmark& \checkmark&           & \checkmark&           & 11.5 & 11.3 \\
\checkmark& \checkmark& \checkmark& \checkmark&           & 13.1 & \textbf{14.3} \\
\rowcolor{gray!10}
\checkmark& \checkmark& \checkmark& \checkmark& \checkmark & \textbf{13.9} & 14.1 \\
\bottomrule
\end{tabular}
\end{table}

\begin{table}[t]
\caption{Ablation study on background‐interference suppression.}
\label{ablation_BS}
\centering
\small
\setlength{\tabcolsep}{5pt}
\begin{tabular}{@{}ccccc cc@{}}
\toprule
% \multicolumn{4}{c}{\textbf{Suppress Settings}} &
% \multicolumn{2}{c}{\textbf{mAP (\%)}} \\
\cmidrule(lr){1-4}\cmidrule(lr){5-6}
\multicolumn{2}{c}{\textbf{w/ suppress}} &
\multicolumn{2}{c}{\textbf{w/o suppress}} &
\textbf{GZSD} & \textbf{ZSD} \\
\cmidrule(lr){1-2}\cmidrule(lr){3-4}\cmidrule(lr){5-6}
TG-FE & TG-QE & TG-FE & TG-QE & HM & AP$_{50}$ \\
\midrule
           &            &            &            &  8.8 &  9.8 \\
           &            & \checkmark &            &  9.3 & 11.7 \\
\checkmark &            &            &            & 10.4 & 12.3 \\
\checkmark &            &            & \checkmark & 11.6 & 12.5 \\
\checkmark & \checkmark &            &            & 13.1 & 14.3 \\
\bottomrule
\end{tabular}
\end{table}

% These results indicate that the proposed text-guided strategy enable the model to focus on class-relevant features and enhance image-text alignment. 

\noindent \textbf{Ablation Study on Interference Suppression.}
Table~\ref{ablation_BS} demonstrates the effectiveness of the interference suppression branch. Experiments show that using text information as guidance to enhance features in either the encoder or decoder is effective. Compared to the baseline, this approach yields improvements of 0.5\% and 1.2\% in HM of AP$_{50}$, respectively. Introducing the suppression branch further improves detection accuracy. For instance, for TG-QE module, introducing suppression branch further improves the HM of AP$_{50}$ by 1.5\%.

\begin{figure}[!t]
\centering
\includegraphics[width=\linewidth]{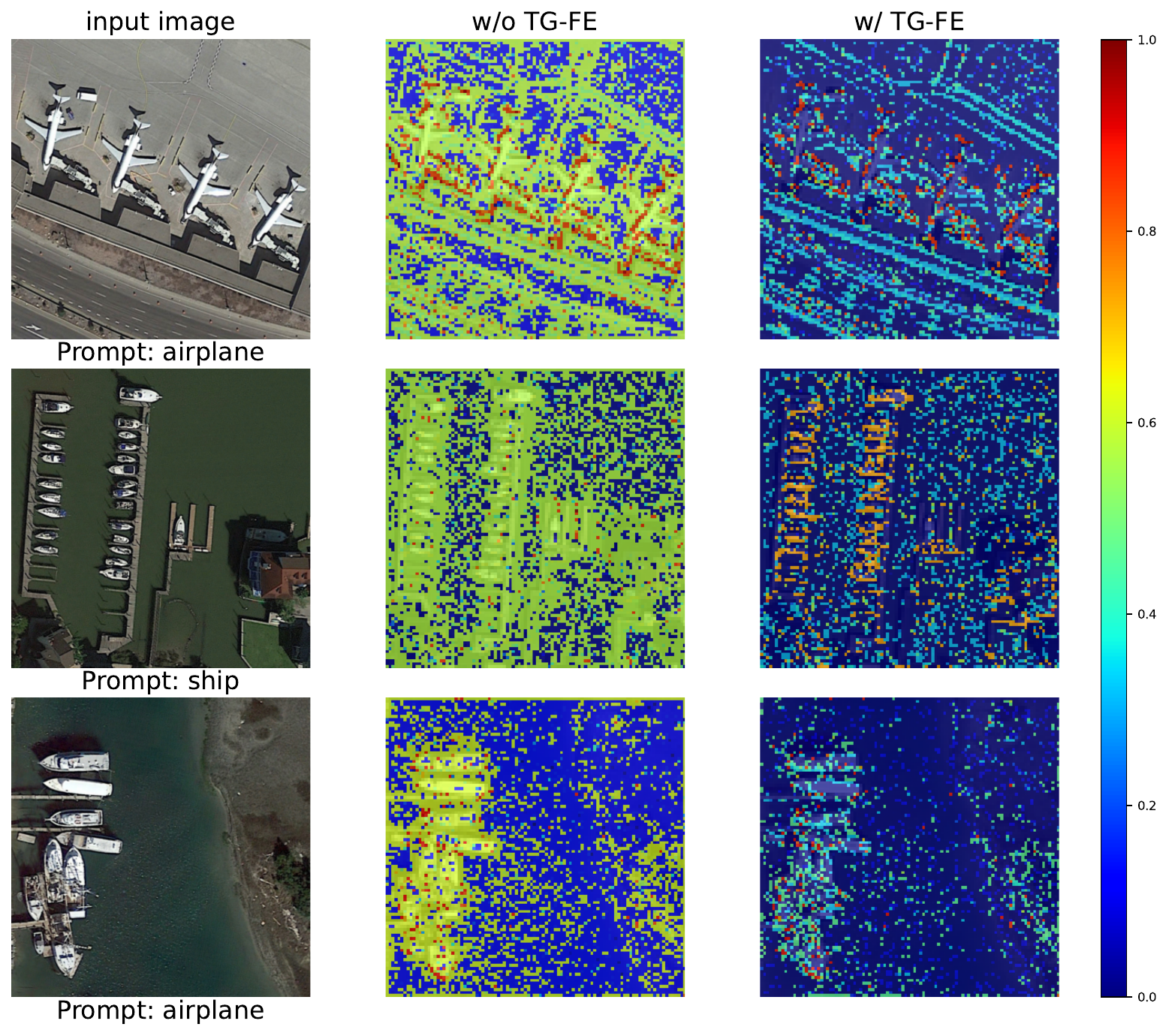}
\caption{The impact of the TG-FE module on features. The first column displays the input images, while the subsequent two columns illustrate the correlation between the image features and the detected targets.}
\label{fig:TG-FE-impact}
\end{figure}

\noindent \textbf{Visual Analysis on the Impact of the proposed TG-FE.} Fig.\ref{fig:TG-FE-impact} visualize the impact of the TG-FE module on the features. Specifically, we visualized the correlation between visual features and the text prompt before and after the introduction of the TG-FE module. The first row corresponds to cases with larger objects, while the second row corresponds to cases with smaller objects. Comparing the results in the second and third columns, it is evident that after applying the TG-FE module, the correlation between visual features and the text prompt is significantly enhanced. In particular, in the second-row example where the ship is relatively small, the initial visual features exhibited a vague correlation spread across various parts of the image. After applying TG-FE, the ship's features became much more prominent, and the background was effectively suppressed. As shown in the third row, if the text prompt does not match the image content, the entire image is suppressed. This result, viewed from a visualization perspective, demonstrates the effectiveness of the proposed TG-FE. 
VG-TR and TG-QE is analogous to TG-FE. TG-FE and VG-TR operates on the encoder, and the TG-QE on the decoder; together, they work synergistically to ensure the precise detection of RT-OVAD.

\section{Conclusion}
In this paper, we propose RT-OVAD, aiming to develop an efficient open-vocabulary detector for aerial object detection. To this end, we focus on exploring image-text collaboration. Specifically, we incorporate class semantics into the detector and introduce a image-to-text alignment loss to enhance image-text alignment, thereby eliminating the predefined category constraints of traditional detectors. We further propose TG-FE, VG-TR and TG-QE, forming a image-text collaboration encoder-decoder framework. This architecture makes the model focus on class-relevant features, allowing it to utilize rich textual information for improved detection performance. Moreover, by avoiding computationally intensive modules, RT-OVAD inherits the efficiency of its baseline, meets the real time demand of aerial detection, and is therefore more suitable for open scenarios. In future work, we will further explore text guidance, extending word-level cues to phrases and even sentences to provide more detailed guidance information, thereby enhancing open detection capabilities while broadening the model's applicability.

% \clearpage
% \setcounter{page}{1}
% \setcounter{figure}{0}
% \setcounter{section}{0}
% \setcounter{table}{0}
% \maketitlesupplementary

\begin{figure*}[t!]
\centering
\includegraphics[width=0.8\linewidth]{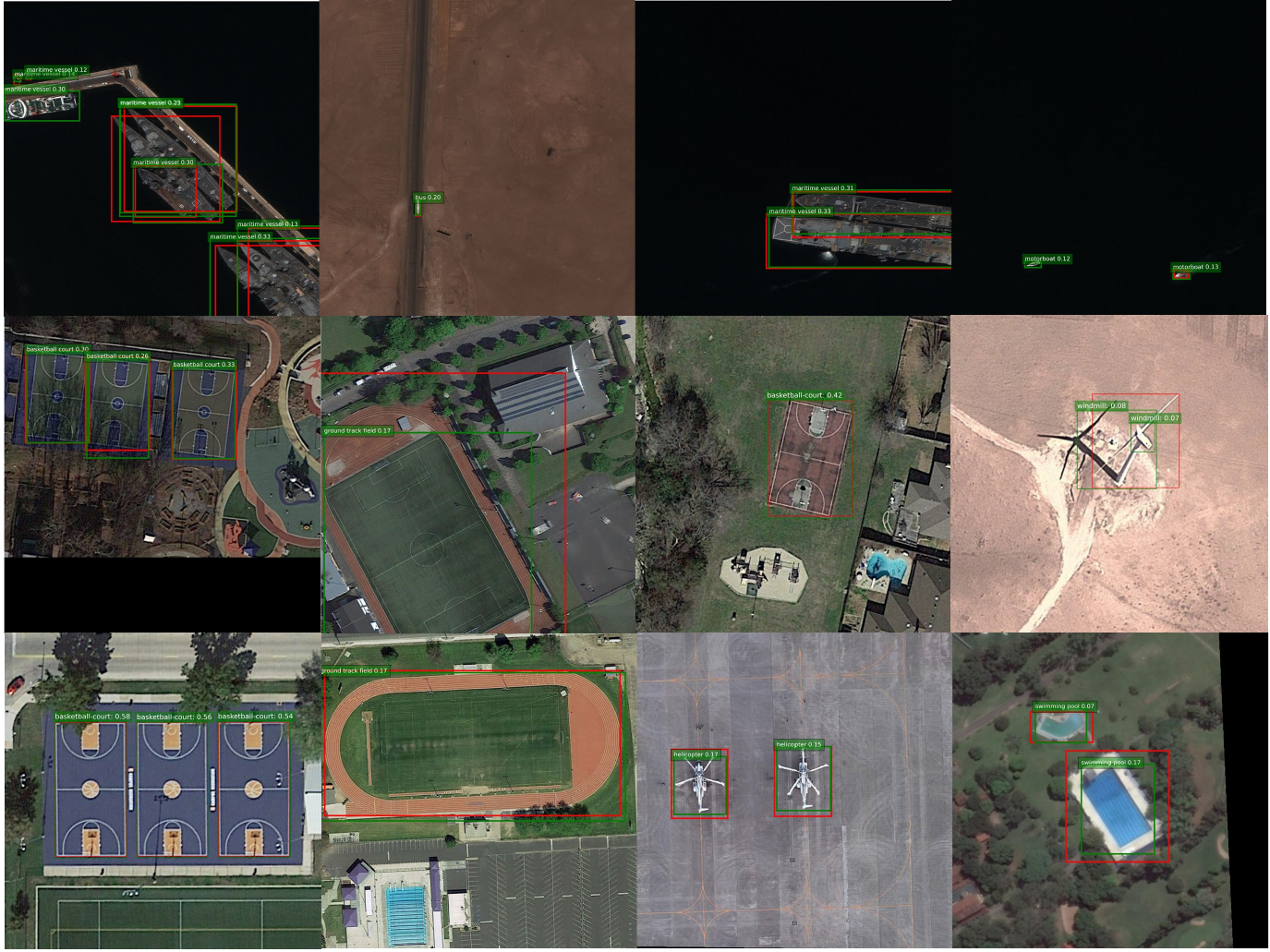} % Reduce the figure size so that it is slightly narrower than the column.
\caption{Qualitative results for zero-shot detection on the xView, DIOR, and DOTA datasets, focusing on novel classes. The green rectangles represent predicted bounding boxes, while red rectangles denote ground truth bounding boxes.}
\label{visualization}
\end{figure*}

\appendix
\section{Dataset Details}
\subsection{Zero-shot Detection Dataset Details}
To ensure a fair experimental comparison, we followed the state-of-the-art DescReg \cite{zang2024zero} for open aerial object detection to organize our experiments. We evaluate RT-OVAD zero-shot detection performance on three benchmark datasets. The dataset details as follows:
\begin{enumerate}
    \renewcommand{\labelenumi}{\textbullet} 
    \item \textbf{xView} \cite{lam2018xview} comprises high resolution satellite images captured by the WorldView-3 satellite at a 0.3m ground sample distance. The resolutions in xView range from about \(2500 \times 2500\) to \(5000 \times 3500\) pixels. The dataset includes 846 annotated images with 60 classes. We selected 665 images for training and 181 for evaluation.
    \item \textbf{DIOR} \cite{li2020object} is a large-scale benchmark dataset for aerial object detection, with the resolution of \(800 \times 800\). It consists of 5862 training images, 5863 validation images, and 11,725 test images, with 20 annotated classes. We conducted training on the training set and evaluation on the validation set.
    \item \textbf{DOTA} \cite{xia2018dota} consists of satellite images with resolutions ranging from \(800 \times 800\) to \(4000 \times 4000\). In this work, we employ both DOTAv1.0 and DOTAv1.5, which share the same set of 1411 training images and 458 validation images, but differ in that DOTAv1.5 includes additional annotations for very small objects.
\end{enumerate}

\subsubsection{Dataset Crop Details}
For the xView and DOTA datasets, due to their high image resolutions, we simplified the datasets by cropping images during pre-processing. Following prior work \cite{zang2024zero}, we cropped the images in these datasets to a size of \(800 \times 800\). Ultimately, we obtained 12,825 training images and 3,550 test images for xView, and 18,430 training and 6,259 test images for DOTA, respectively. 

% Due to the unavailability of the DescReg dataset split, we independently divided the dataset according to the descriptions provided in its publication; our splits for the xView dataset slightly differ from the described 12,826 training and 3,549 test images.

\subsubsection{Dataset Base/Novel Split Details} 
For the xView and DIOR datasets, we followed the setup from prior work \cite{zang2024zero} to split base and novel classes. The resulting category splits are 48/12 and 16/4 for xView and DIOR, respectively.
The class splits are:
\begin{enumerate}
    \renewcommand{\labelenumi}{\textbullet} 
    \item \textbf{xView dataset}:  
    \begin{enumerate}
        \renewcommand{\labelenumii}{$\circ$}
        \item Base: 'fixed wing aircraft', 'small aircraft', 'passenger plane or cargo plane', 'passenger vehicle', 'small car', 'utility truck', 'truck', 'cargo truck', 'truck tractor', 'trailer', 'truck tractor with flatbed trailer', 'truck tractor with liquid tank', 'crane truck', 'railway vehicle', 'passenger car', 'cargo car or container car', 'flat car', 'tank car', 'locomotive', 'sailboat', 'tugboat', 'fishing vessel', 'ferry', 'yacht', 'container ship', 'oil tanker', 'engineering vehicle', 'tower crane', 'container crane', 'straddle carrier', 'dump truck', 'haul truck', 'front loader or bulldozer', 'cement mixer', 'ground grader', 'hut or tent', 'shed', 'building', 'aircraft hangar', 'damaged building', 'facility', 'construction site', 'vehicle lot', 'helipad', 'storage tank', 'shipping container', 'pylon', 'tower'.
        \item Novel: 'helicopter', 'bus', 'pickup truck', 'truck tractor with box trailer', 'maritime vessel', 'motorboat', 'barge', 'reach stacker', 'mobile crane', 'scraper or tractor', 'excavator', 'shipping container lot'.
    \end{enumerate}
    \item \textbf{DIOR dataset}: 
    \begin{enumerate}
        \renewcommand{\labelenumii}{$\circ$}
        \item Base: 'airplane', 'baseball field', 'bridge', 'chimney', 'dam', 'Expressway Service area', 'Expressway toll station', 'goldfield', 'harbor', 'overpass', 'ship', 'stadium', 'storage tank', 'tennis court', 'train station', 'vehicle'.
        \item Novel: 'airport', 'basketball court', 'ground track field', 'windmill'.
    \end{enumerate}
\end{enumerate}
For the DOTA dataset, we observed overlaps between the novel categories in DIOR and the base categories in prior DOTA splits. In addition, DOTAv1.0 lacks annotations for many small instances. Therefore, we adopt DOTAv1.5 and adjust the category splits to prevent novel class leakage and reduce annotation inconsistencies. As a result, the revised splits yield 12 base classes and 4 novel classes for the DOTA dataset.

\begin{enumerate}
    \renewcommand{\labelenumi}{\textbullet} 
    \item \textbf{DOTA dataset}:  
    \begin{enumerate}
        \renewcommand{\labelenumii}{$\circ$}
        \item Base: 'plane', 'ship', 'storage tank', 'baseball diamond', 'basketball court', 'ground track field', 'harbor', 'bridge', 'large vehicle', 'small vehicle', 'roundabout'.
        \item Novel: 'tennis court', 'helicopter', 'soccer ball field', 'swimming pool'.
    \end{enumerate}
    \item \textbf{DOTA dataset resplit}: 
    \begin{enumerate}
        \renewcommand{\labelenumii}{$\circ$}
        \item Base: 'plane', 'ship', 'storage tank', 'baseball diamond', 'tennis court', 'soccer ball field', 'harbor', 'bridge', 'large vehicle', 'small vehicle', 'roundabout', 'container crane'.
        \item Novel: 'basketball court', 'helicopter', 'ground track field', 'swimming pool'.
    \end{enumerate}
\end{enumerate}

% \subsection{Traditional Detection Dataset Details}
% \begin{enumerate}
%     \renewcommand{\labelenumi}{\textbullet}
%     \item \textbf{Visdrone} \cite{du2019visdrone} is a widely-used benchmark dataset for aerial object detection. The image resolutions vary from \(960 \times 540\) to \(2000 \times 1500\). The dataset comprises 10 categories and consists of 6471 training images, 548 validation images, and 3190 test images. Following prior work \cite{du2023adaptive}, we conducted our evaluation on the validation data.
%     \item \textbf{UAVDT} \cite{du2018unmanned} is also a benchmark for evaluation in aerial object detection. The dataset comprises 23,258 training images and 15,069 testing images with a resolution of \(1024 \times 540\), covering 3 classes.
% \end{enumerate}

\section{Qualitative Results}
As shown in Fig~\ref{visualization}, we present qualitative results of RT-OVAD on the xView, DIOR, and DOTA datasets. The images in the first, second, and third rows correspond to xView, DIOR, and DOTA, respectively. 
These results demonstrate that our proposed method breaks the category limitations of traditional detectors and can accurately detect novel classes.
As can be seen, even for targets with inconspicuous visual features, such as the bus and motorboat in the first row, our proposed RT-OVAD can still detect them.

\section{Implementation Details}
\subsection{Training Details}
The RT-OVAD implementation is based on RT-DETR~\cite{zhao2024detrs} with a ResNet50~\cite{he2016deep} backbone, using the default hyperparameters provided by RT-DETR. In the query selection module, we replace classification scores with image-text similarity scores to select the top 500 encoder features, which are then used to initialize the object queries for the decoder. We also use these similarity scores instead of classification scores for label assignment. For open-vocabulary aerial detection, following the protocol in~\cite{pan2025locate}, all models are first pretrained on the large-scale LAE-1M dataset and subsequently evaluated on the DIOR, DOTAv2.0, and xView benchmarks. In the zero-shot setting, we adopt a one-stage training strategy, where novel categories are excluded from the training sets of all three datasets. The models are evaluated on both generalized zero-shot detection (GZSD) and zero-shot detection (ZSD) tasks using the validation sets of the corresponding benchmarks.

Following the approach in~\cite{li2022grounded, cheng2024yolo}, we adopt a random class sampling procedure during training. Specifically, we shuffle the order of input text each time, preventing the model from relying on fixed positions of class embeddings. We first sample the positive classes that appear in the images, and then randomly sample the remaining negative classes to reach the required number of training classes. If there are insufficient negative classes, we use an empty token as a placeholder and mask the corresponding class embeddings in subsequent operations.

\subsection{Details of the compared method}
% Due to the unavailability of the code from prior work \cite{zang2024zero}, we directly use the results reported in their publication for experiments under their dataset split. YOLO-World~\cite{cheng2024yolo} and Grounding DINO~\cite{liu2024grounding} are two effective OVD methods for natural images. We compare our model against these methods to demonstrate its suitability for aerial OVD tasks. Specifically, we follow the training procedures detailed in their published papers and available code, conducting experiments under settings similar to ours. 
For open-vocabulary aerial detection, we compare RT-OVAD with four representative open-vocabulary detectors, including both general-purpose~\cite{li2022grounded, cheng2024yolo, liu2024grounding} and aerial image–tailored models~\cite{pan2025locate}. Except for RT-OVAD and YOLO-World, all results are directly taken from the LAE-DINO~\cite{pan2025locate} paper for consistency and fairness.

For zero-shot aerial detection, Due to the unavailability of the code from prior work \cite{zang2024zero}, we directly use the results reported in their publication for experiments under their dataset split. We also compare our model against the YOLO-World~\cite{cheng2024yolo}, Grounding DINO~\cite{liu2024grounding} and LAE-DINO~\cite{pan2025locate} to demonstrate the RT-OVAD's suitability for aerial zero-shot detection tasks. Specifically, we follow the training procedures detailed in their published paper and available code, conducting experiments under settings similar to ours.

\subsection{{Limitation Analysis}}
While RT-OVAD achieves strong performance across three prevalent benchmark tasks, several limitations persist:
(1) Limited textual annotations. Current open-vocabulary aerial benchmarks provide only category-level labels. The lack of phrase- or sentence-level descriptions limits the richness and expressiveness of textual guidance during both training and inference.
(2) Potential instability during zero-shot evaluation. 
Under zero-shot settings, we observe that the model exhibits mild performance fluctuations. We attribute these variations to: (i) the sparsity of textual prompts for novel classes, (ii) the long-tailed distribution common in aerial categories, and (iii) inherent stochasticity in the training and inference processes. To improve evaluation stability, we report the average results over three independent runs.
% \textbf{(3) High recall but low precision on novel classes.} Under certain base/novel split scenario, we observe many false positives exsiting, where background regions are incorrectly classified as novel objects. This may indicates that the image-to-text alignment loss lacks an explicit constraint for background regions, resulting in the model's difficulty to accurately distinguish novel objects from complex background features.
(3) High recall but low precision on novel classes. In certain base/novel class split settings, we observe a substantial number of false positives, where background regions are mistakenly classified as novel objects. This may indicate that the image-to-text alignment loss lacks explicit constraints for background regions, making it difficult for the model to accurately distinguish novel targets from complex background features.

Despite these limitations, RT-OVAD still delivers state-of-the-art accuracy while maintaining real-time inference speed. Future work may consider: (i) enriching benchmark datasets with fine-grained textual annotations, and (ii) exploring prompt-learning strategies or background-aware loss functions to improve model discrimination.

\bibliographystyle{IEEEtran}
\bibliography{OVA_Det}

\end{document}